\documentclass[journal]{IEEEtran}

\usepackage{graphicx}
\graphicspath{ {./Fig/} }
\usepackage[caption=false,font=normalsize,labelfont=sf,textfont=sf]{subfig}
\usepackage{textcomp}
\usepackage{stfloats}
\usepackage{physics}
\usepackage{amsmath, amsfonts}
\usepackage{xcolor, soul}
\usepackage[hidelinks]{hyperref}
\usepackage{multirow}
\usepackage{cite}

\usepackage{soulutf8}
\DeclareMathAlphabet{\mathpzc}{OT1}{pzc}{m}{it}
\newcommand{\state}{\mathpzc{x}}

\DeclareMathOperator*{\minimize}{\text{minimize}}
\DeclareMathOperator*{\subjto}{\text{subject to}}

\DeclareMathOperator*{\argmin}{arg\,min}

\newcommand{\intd}{\mathrm{d}}

\newcommand{\traj}{r_\mathrm{d}}

\newcommand{\changes}[1]{#1}
\newcommand{\TR}[1]{#1}

\title{\changes{Autonomous Hook-Based Grasping and Transportation with Quadcopters}}
\author{Péter Antal, Tamás Péni, and Roland Tóth
\thanks{This research was supported by the European Union within the framework of the National Laboratory for Autonomous Systems (RRF-2.3.1-21-2022-00002) and by the Hungarian Research Network (grant. number: SA-77/2021).}
\thanks{The authors are with the Systems and Control Lab, HUN-REN Institute for Computer Science and Control, Budapest, Hungary (email: \{antalpeter,peni,tothroland\}@sztaki.hun-ren.hu). R. T\'{o}th is also affiliated with the Control Systems Group of the Eindhoven University of Technology, The Netherlands.}}

\begin{document}

\maketitle

\begin{abstract}
\changes{Payload grasping and transportation with quadcopters is an active research area that has rapidly developed over the last decade. To grasp a payload without human interaction, most state-of-the-art approaches apply robotic arms that are attached to the quadcopter body. However, due to the large weight and power consumption of these aerial manipulators, their agility and flight time are limited. This paper proposes \TR{a motion control and planning method for transportation with a} lightweight, passive manipulator structure that consists of a hook attached to a quadrotor using a 1 DoF revolute joint.} To perform payload grasping, transportation, and release, first, time-optimal reference trajectories are designed through specific waypoints to ensure the fast and reliable execution of the tasks. Then, a two-stage motion control approach is developed based on a robust geometric controller for precise and reliable reference tracking and a linear--quadratic payload regulator for rapid setpoint stabilization of the payload swing. \changes{
\TR{Furthermore,} stability of the closed-loop system i\TR{s mathematically proven to give safety guarantee for its operation.} 
The proposed control architecture and design are evaluated in a high-fidelity physical simulator, and also in real flight experiments, using a custom-made quadrotor--hook manipulator platform.}
\end{abstract}


\section{Introduction}

\changes{Due to their advanced actuators, sensors, and autonomous navigation capability, 
\TR{interest for}
quadcopters is continuously growing both in industry and research. Nowadays, most industrial applications of quadrotors \TR{are limited to} passive tasks, such as agricultural monitoring or recording camera footages. However, as drone technology matures, more and more challenges arise which require direct \TR{interaction and active manipulation of the environment}.} Our work is motivated by current industrial developments, where drones are used to solve autonomous transportation tasks, e.g. transporting parts between production cells \cite{Maghazei2020}. \changes{In these applications, the grasping and release requires a gripping mechanism that can connect the payload to the drone, preferably without human interaction. Moreover, an advanced control system has to be developed to safely and reliably execute the task.}

In the literature of transportation with quadcopters, the payload is usually connected to the vehicle via flexible cables that are attached manually before the drone takes off \cite{Gassner2017,Hua2021,Li2021,li_autotrans_2023}. The desired payload motion is then either achieved by swing-free trajectory planning and control \cite{alkomy_vibration_2021, guerrero-sanchez_swing-attenuation_2017}, or exploiting differential flatness of the coupled system to design dynamically feasible, smooth trajectories
\cite{Yu2020,Sreenath2013}. Advantages of cable suspension include good maneuverability, simple mechanical structure, and the ability to apply multiple vehicles to collaboratively transport a heavy payload. \changes{However, due to the large number of degrees of freedom, object grasping without human interaction is not yet solved.}

\begin{figure}
\centering 
\includegraphics[width=\linewidth]{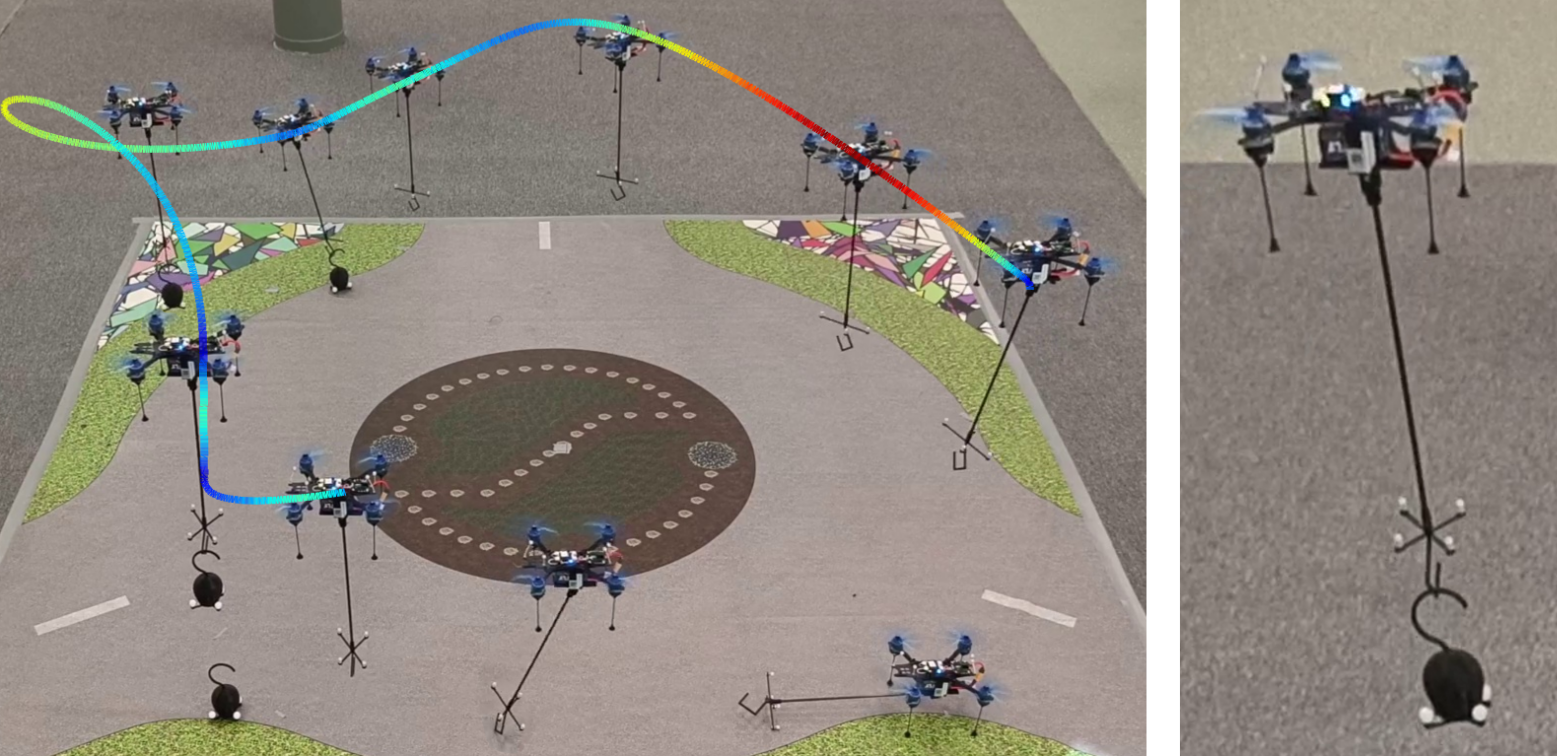}
\caption{Payload transportation by a hook-based manipulator.}\label{fig:real_manipulator} 
\end{figure}

\changes{For automatic object grasping and transportation, quadcopters can be equipped with robotic extensions 
\TR{for} aerial \TR{manipulations} \cite{Ruggiero2018, Ollero2022,saunders_autonomous_2024}. In \cite{Thomas2014}, autonomous grasping is proposed with robotic claws, inspired by how predatory birds grasp their prey. The modelling and control design of the aerial manipulator are simplified by considering only the dynamics in the vertical plane. 
In \cite{Augugliaro2014}, a custom-designed gripper is proposed specifically for grasping and carrying foam elements. In \cite{zhang_grasp_2018}, a hex-rotor equipped with a 7 DoF manipulator is applied to grasp a moving payload. To control the motion of the multirotor, a PID controller has been developed. In \cite{luo_gaussian_2022}, a 1 DoF magnetic gripper is attached to the quadcopter, and minimum snap trajectories are generated between predefined waypoints to complete the payload grasping and transportation task. Using the 1 DoF manipulator, the authors proposed a minimal time trajectory planning method in \cite{luo_time-optimal_2023} to grasp a moving object. Although the algorithm is able to generate dynamically feasible time-optimal trajectories, the computation time is over 10 seconds, therefore real-time replanning is not possible.

Compared to cable suspension, the main advantage of robotic arms and grippers is that they are actuated, which makes gripping possible even in the presence of small positioning errors. However, most of them are either designed for a specific task, material and payload, or require large and less agile quadcopter platforms, which, due to their heavy weight and power consumption, limit general applicability.}

\changes{In this work, we combine the agility and maneuverability of cable suspension with the automatic grasping and transportation ability of aerial manipulators.} We propose a grasping mechanism that consists of a hook fixed to a rigid rod, which is connected to a quadrotor by a 1 DoF passive revolute joint. Using this quadrotor--hook manipulator, 
we intend to move objects (with a hook attached on top) by picking them up and putting them down precisely at pre-specified locations within minimal time. 
The advantages of our setup compared to the previously introduced configurations are as follows. First, the applied pole is more rigid than flexible cables and only has one degree of freedom, which decreases the complexity of precise positioning and makes modelling and tracking of the cable tension unnecessary. Second, the gripping mechanism is not actuated, resulting in a simple, lightweight and cheap mechanical design. Moreover, the solution is also scalable: by equipping a heavy payload with hooks, it can be transported using multiple vehicles collaboratively. 

\changes{The operational advantages brought by the proposed gripper mechanism come with the price of major unresolved challenges in control system design. To attach and detach the hooks, advanced planning and precise positioning are required, because large tracking errors can lead to unsuccessful grasping and release of the payload. Moreover, the control system of the quadcopter needs to handle the force and torque disturbances introduced by the hook mechanism with attached load. Finally, many existing methods, such as differential flatness-based algorithms, cannot be applied to 1 DoF hook-based manipulators, and there are few works that address the complete task of payload grasping, transportation and release.}



To address the introduced challenges, the contributions of our work are as follows:
\begin{enumerate}
    \changes{\item We propose a path planning solution adapted to hook-based transportation: the reference trajectory is constructed based on 5 segments, each designed using a computationally efficient, minimal time planning method inspired by \cite{Verscheure2009}.
    \item We derive an extension of the robust geometric controller originally suggested in \cite{Lee2013} to achieve precise and reliable tracking of agile trajectories despite all the disturbances caused by the hook and the payload.
    \item For rapidly damping the swing of the payload and ensuring precise release at the target location, \TR{we switch to} a \emph{linear--quadratic payload regulator} (LQR). We mathematically prove that the closed-loop system remains stable despite of the switching between the geometric controller and the LQR.}
    \item We test and analyze the performance of the proposed method in a high-fidelity simulation environment and also in real-world flight experiments using a custom-made 
    aerial manipulator platform.
\end{enumerate}

\changes{To the best of our knowledge, we provide the first solution for autonomous payload grasping and transportation with a fully passive manipulator attached to a quadcopter.

This work is organized as follows. First, we introduce
the dynamic model of the quadrotor and the hook-based manipulator in Section~\ref{sec:model}. In Sections~\ref{sec:traj} and \ref{sec:control},  the proposed trajectory planning and control algorithms are presented. In Sections~\ref{sec:simulation} and \ref{sec:exp}, we evaluate the performance of the proposed methods in high-fidelity physical simulations, and also in real-world flight experiments. Finally, the conclusions are presented in Section~\ref{sec:summary}.}

\section{Modelling}\label{sec:model}

\subsection{Quadrotor Dynamics}\label{sec:quad_dyn}
\begin{figure}[t]

    \centering 
    \includegraphics[height=3.5cm]{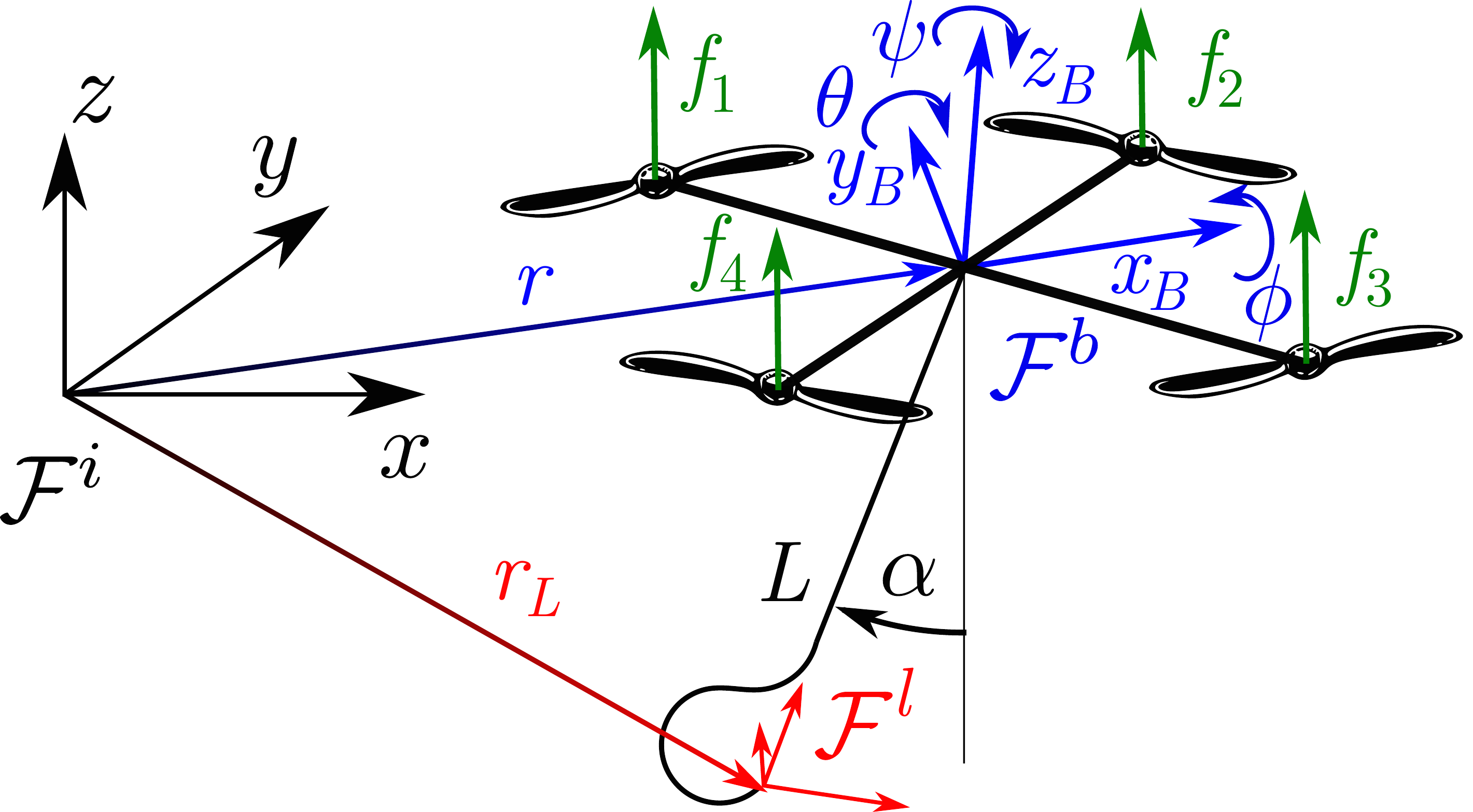}
    \caption{Coordinate frames describing the geometric relations of the quadcopter and the environment.}\label{fig:quad_model}
\end{figure}

In this subsection, the dynamic model of the quadrotor without the hook is derived. The vehicle is modelled as a rigid body in 3D space affected by the gravitational force and the rotation of the propellers. Using the inertial frame ($\mathcal{F}^\mathrm{i}$) and the body frame ($\mathcal{F}^\mathrm{b}$) illustrated in Fig.~\ref{fig:quad_model}, the motion of the quadcopter is governed by the following equations:
\begin{subequations}\label{eq:quad_dyn}
\begin{align}
    m\ddot{r} &= \underbrace{-m g e_3 + F R e_3}_{f_\mathrm{quad}^\mathrm{r}},\label{eq:quad_dyn_1}\\
    J\dot{\omega} &= \underbrace{\tau - \omega \times J \omega}_{f_\mathrm{quad}^\omega},\label{eq:quad_dyn_2}\quad \dot{R} = R \hat{\omega},
\end{align}
\end{subequations}
where $r(t)\in\mathbb{R}^3$ is the position of the drone, $R(t)\in\mathrm{SO}(3)$ is the rotation matrix from $\mathcal{F}^\mathrm{b}$ to $\mathcal{F}^\mathrm{i}$, SO(3) denotes the special orthogonal group, 
$\omega(t)\in\mathbb{R}^3$ is the angular velocity in $\mathcal{F}^\mathrm{b}$, $m$ is the mass of the drone, $J$ is the inertia matrix in $\mathcal{F}^\mathrm{b}$, $g$ is the gravitational acceleration, and $e_3$ is the unit vector of the $z$ axis in $\mathcal{F}^\mathrm{i}$. The notation $\hat{\cdot}$ stands for the projection: $\mathbb{R}^3\rightarrow \mathrm{SO}(3)$ ensuring that $\hat{x}y = x\times y$ for all $x,y\in \mathbb{R}^3$ where $\times$ corresponds to the vector product. Beside of $R$, we also use the Euler angles (roll: $\phi$, pitch: $\theta$, yaw: $\psi$) to characterize the rotation from $\mathcal{F}^\mathrm{b}$ to $\mathcal{F}^\mathrm{i}$, as it is illustrated in Fig.~\ref{fig:quad_model}. The inputs of the system are the collective thrust $F$ and the torque vector $\tau = [\ \tau_\mathrm x\ \tau_\mathrm y\ \tau_\mathrm z\ ]^\top$ \cite{Mellinger2011}.

The dynamic model of the quadcopter is underactuated as it has 6 DoFs, but only 4 control inputs. However, in trajectory design, we exploit that the model is \textit{differentially flat} \cite{Nieuwstadt1998,Mellinger2011}, i.e., all state variables and control inputs can be expressed from a finite number of time derivatives of the \textit{flat outputs}, namely the position $x,y,z$ and yaw angle $\psi$. This simplifies the trajectory planning as all other states depend on the desired $x_\mathrm d(t), y_\mathrm d(t), z_\mathrm d(t),$ $\psi_\mathrm d(t)$ trajectories, therefore only these have to be constructed instead of considering the full state. Moreover, utilizing differential flatness resolves underactuation, because the reference trajectories designed for the flat outputs are inherently dynamically feasible.
\vspace{-1mm}

\subsection{Manipulator Dynamics}

Next we complete the quadcopter model with the manipulator dynamics based on the modelling concept in \cite{Lippiello2012}. We choose the vector of generalized coordinates as $\xi = [\ r^\top\ \lambda^\top \ \alpha\ ]^\top,$ $\xi(t) \in \mathbb{R}^7$, where $\lambda = [\ \phi\ \theta\ \psi\ ]^\top$ and $\alpha(t) \in \mathbb{R}$ is the swing angle, shown in Fig.~\ref{fig:quad_model}. Then, we use the Euler-Lagrange formulation to construct the equations of motion. For this, we introduce the Lagrange function $\mathcal L = \mathcal K - \mathcal U$, where $\mathcal K$ and $\mathcal U$ are the kinetic and potential energy of the system, computed as follows:
\begin{subequations}
    \begin{align}
    \mathcal K = & \frac{1}{2} m \dot{r}^\top \dot{r} + \frac{1}{2} \dot{\lambda}^\top Q^\top J Q \dot\lambda +  \frac{1}{2} m_\mathrm L  \dot{r}_\mathrm{L}^\top \dot{r}_\mathrm{L},\label{eq:lagr_fun}\\ 
    \mathcal U = &\ m g r^\top \! e_3 + m_\mathrm{L} g (r + R r_\mathrm b^\mathrm l)^\top e_3,
\end{align}
\end{subequations}
where $\dot{r}_\mathrm{L} = \dot{r} - \widehat{R r_\mathrm b^\mathrm l} R Q \dot\lambda + R \dot r _\mathrm b^\mathrm l$, $Q(t) \in \mathbb{R}^{3\times 3}$ is the standard transformation matrix from the derivatives of the Euler angles to the angular velocity \cite{Lippiello2012}, $L$ is the pole length, $r_\mathrm b^\mathrm l = [\ -L\!\sin(\alpha) \hspace{2mm} 0  \hspace{1mm} -\!\!L\cos(\alpha) \ ]^\top$, and $m_\mathrm{L}$ is the payload mass. Using the Euler-Lagrange equations $\dv{}{t}\frac{\partial \mathcal L}{\partial \dot{\xi_i}} - \frac{\partial \mathcal L}{\partial \xi_i} = u_{\mathrm{g}, i}$ for $i=1\dots 7$, the dynamics of the manipulator are
\begin{align}
    H(\xi)\ddot{\xi} + C(\xi, \dot\xi) \dot\xi + G(\xi) = \Xi(\xi) u,\label{eq:full_dyn}
\end{align}
with coefficient matrices 
    \begin{align}
    \begin{split}
           &H_{i, j} = \frac{\partial^2 \mathcal K}{\partial \dot\xi_i \partial \dot\xi_j},\quad G_k = \frac{\partial \mathcal U}{\partial \xi_k},\quad \Xi = \begin{bmatrix}
        R e_3 & 0_{3\times 3} \\ 0_{3\times 1} & Q^\top \\ 0_{3\times 1} & 0_{3\times 3}
    \end{bmatrix},\\
    &C_{i, j} = \sum_{k=1}^{7} \frac{1}{2} \left( \frac{\partial H_{i, j}}{\partial \xi_k} +  \frac{\partial H_{i, k}}{\partial \xi_j} +  \frac{\partial H_{j, k}}{\partial \xi_i}  \right) \dot{\xi}_k
    \end{split}
\end{align}
for $i, j, k \in \{1,\dots , 7\}$. The control input vector is $u = [\ F\ \tau^\top \ ]^\top$, and $u_{\mathrm{g}} = \Xi(\xi) u$ is the generalized force vector. From \eqref{eq:full_dyn}, $\ddot\xi = f_\mathrm{full}(\xi, \dot\xi, u)$ can be expressed, which we will use later in controller design. 

Despite of singularity of \eqref{eq:full_dyn} due to the use of Euler angles, this model is adequate for the control problem at hand where we will only use it to describe motions of the payload around the stable equilibrium point. Compared to the quadrotor dynamics, the hook mechanism with attached load introduces zero moment in pitch, a non-zero moment in roll and yaw, and a force in the $x-z$ plane.

\section{Motion Trajectory Planning}\label{sec:traj}

\subsection{Overall concept}\label{sec:overview}

\begin{figure}
\centering 
\includegraphics[height=3.1cm]{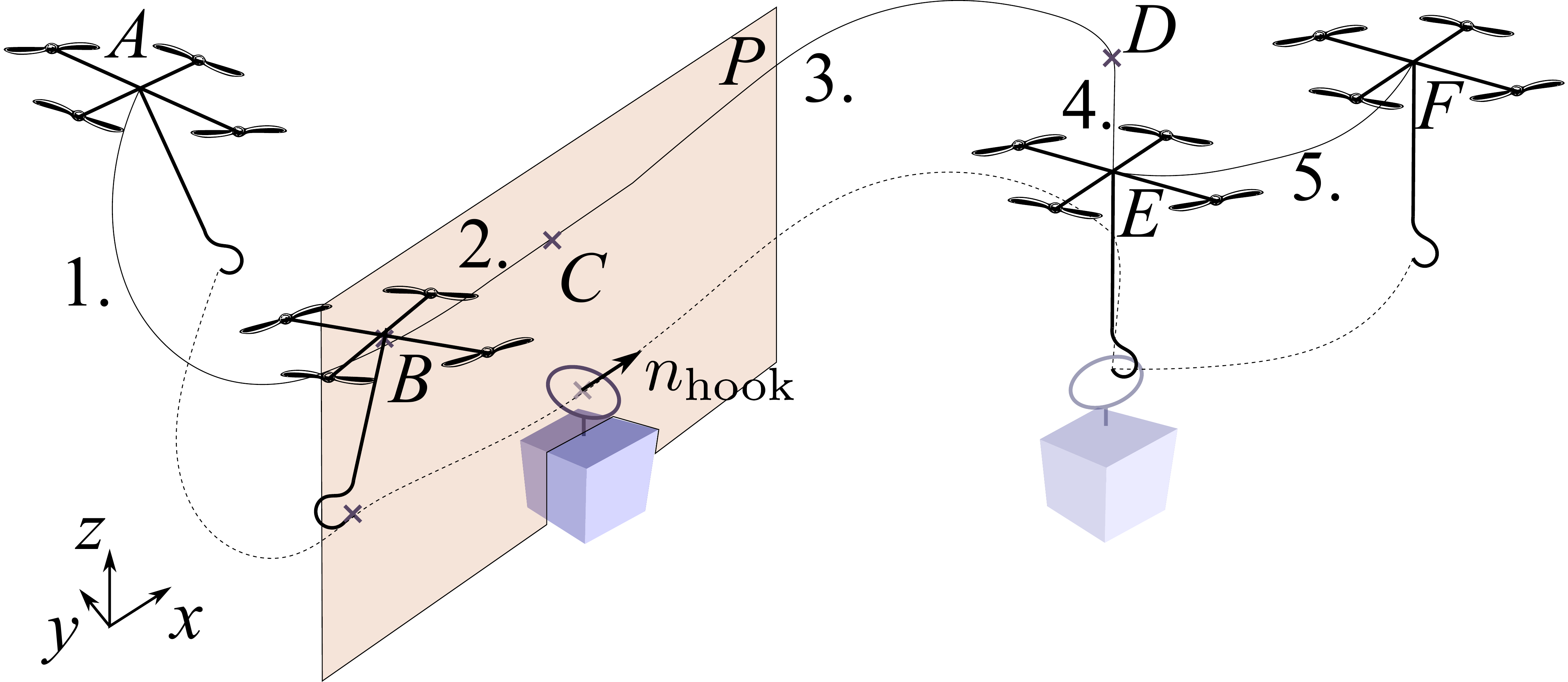}
\caption{Motion trajectory segments and waypoints.}\label{fig:traj} 
\end{figure}

To perform successful payload grasping and transportation, we propose a motion sequence for the aerial manipulator consisting of 5 segments: 1) reach the plane of the payload with the quadrotor; 2) approach the payload and attach the hook; 3) take the payload to the target position; 4) release the payload; 5) detach the hook. The trajectory segments are depicted in Fig.~\ref{fig:traj} with waypoints $A, B, C, D, E, F$ describing the connection points between the segments and the plane $P$, which is spanned by the normal vector of the hook on the payload ($n_\mathrm{hook}$), and the $z$ axis of the world frame. The position of the waypoints $A, C, D, E, F$ are fixed based on the initial and the intended final configuration of the quadrotor and the payload, while the position of waypoint $B$ is determined by the solution of the trajectory optimization.

In trajectory planning, we neglect the dynamics introduced by the manipulator and the payload as their effects will be compensated by the trajectory tracking controller (see  Sec.~\ref{sec:control}). Consequently, we design reference trajectories for the flat outputs of the quadrotor system derived in Sec.~\ref{sec:quad_dyn}, namely the position and yaw angle, while the rest of the states in terms of roll and pitch and their velocities will be completely determined through the kinematic relations. 

\subsection{Spatial (x,y,z) trajectory}
For the position, each segment is designed in two steps, based on an indirect time optimal trajectory planning method inspired by \cite{Verscheure2009}. First, we design a spatial trajectory $\traj(s)$ as a function of the path variable $s$ and then we construct a velocity profile by determining the time dependence of $s$. We represent the spatial trajectory by B-spline curves defined via their basis functions and coefficients as follows:
\begin{equation}
    \traj (s) = \begin{bmatrix} x_\mathrm{d}(s)& y_\mathrm{d}(s)& z_\mathrm{d}(s) \end{bmatrix}^\top =  c^\top\vartheta(s),
\end{equation}
where $\vartheta : \mathbb{R} \rightarrow \mathbb{R}^{n_\mathrm{c}}$ is the B-spline basis of degree~$n_\mathrm{r}$, $c = [\ c_\mathrm{x}\ c_\mathrm{y}\ c_\mathrm{z}\ ]  \in \mathbb{R}^{n_\mathrm{c} \times 3}$ is the coefficient matrix, and $s(t)\in [0, 1]$ is~the progress variable along the B-spline curve. This formulation ensures that the designed path is $n_\mathrm{r}-2$ times continuously differentiable given the knots of the B-splines are distinct. Furthermore, the B-spline and its derivatives are linear in the coefficients allowing us to formulate trajectory optimization as a quadratic problem with linear constraints.

For the spatial trajectory, we minimize the jerk and the arc length of the curves with the objective function \cite{Gao2018}:
\begin{subequations}
\begin{align}
    J_\mathrm{jerk} &= \sum_{f \in \{x_\mathrm{d}, y_\mathrm{d}, z_\mathrm{d}\}} \int_0^{1} \left( \dv[3]{f(s)}{s {}} \right)^{\!\!2} \intd s,\label{eq:obj_snap}\\
    J_\mathrm{length} &= \int_0^{1} \sqrt{\left(x'_\mathrm{d}(s)\right)^2 
    + \left(y'_\mathrm{d}(s)\right)^2 + \left(z'_\mathrm{d}(s)\right)^2} \ \intd s,\\
    J &= w J_\mathrm{jerk} + (1-w) J_\mathrm{length}^2,\quad w \in [0, 1],\label{eq:spat_obj}
\end{align}
\end{subequations}
where $w$ is a trade-off parameter chosen by the user. Throughout this work, $g'(s)$ denotes the derivative of function $g$ \emph{with respect to} (w.r.t.) $s$, and $\dot{g}(s(t))$ denotes the derivative of $g$ w.r.t. time: $\dot{g}(s(t)) =  g'(s(t)) \dot{s}(t)$. 

In order to incorporate the waypoints illustrated in Fig.~\ref{fig:traj}, we prescribe boundary conditions on the spatial trajectory and also on its derivatives. This leads to a quadratic optimization problem of the coefficients of the spatial trajectory:
\begin{align}\label{eq:qp}
\begin{split}
         &\tilde c^* =  \argmin J (\tilde c),\quad \\
    \subjto\ \ & A  \tilde c = b,\quad G \tilde c \leq h,
\end{split}
\end{align}
where $\tilde c = [c^\top_\mathrm{x}\ c^\top_\mathrm{y}\ c^\top_\mathrm{z}]^\top \in \mathbb{R}^{3n_\mathrm{c}}$ and the constraints determined by $A, b, G, h$ are discussed later in Section~\ref{sec:const}.

\subsection{Temporal (x,y,z) trajectory}\label{sec:temp}

The optimal time allocation method we employ has been proposed in \cite{Verscheure2009}, and applied to quadcopter trajectory planning in \cite{Gao2018, Hua2021}. The objective of the planning is to design the velocity profile of $s(t):\mathbb{R}\rightarrow [0, 1]$ such that it minimizes the execution time $T = \int_0^1 \frac{1}{\dot{s}} \ \intd s$ of the trajectory while satisfying motion constraints for the quadrotor. 
Similar to \cite{Gao2018}, we also incorporate the actuation constraints in the objective by penalizing the acceleration along the path, as the control inputs of the quadrotor appear at the acceleration level in the dynamic model \eqref{eq:quad_dyn}. Given that the spatial path $\traj (s)$ is already designed based on \eqref{eq:qp}, the constrained optimal time allocation problem is formulated as follows:
\begin{align}\label{eq:temp_opt}
\begin{split}
  \minimize_{T\in \mathbb{R}_0^+, s\in [0,1]^{[0,T]}} \quad & J_\mathrm{T} = \int_0^1 \left( \rho \ddot{s}^2 + (1-\rho) \frac{1}{\dot{s}} \right)\intd s\\
\subjto \quad & \dot r_\mathrm{d}(0) = v_0, \quad \dot r_\mathrm{d}(1) = v_\mathrm{f}, \\
\quad & -v_\mathrm{max} \leq \dot r_\mathrm{d}(s(t)) \leq v_\mathrm{max},\\
& -a_\mathrm{max} \leq \ddot r_\mathrm{d}(s(t)) \leq a_\mathrm{max},\\
& -\lambda_\mathrm{max} \leq \lambda(s(t)) \leq \lambda_\mathrm{max},\\
& \dot{s}(t) \geq 0, \quad t\in [0, T],
\end{split}
\end{align}
where $\rho \in [0, 1]$ is a tradeoff parameter, and $\lambda(s(t))$ is the rate of the acceleration (similar to jerk, but differentiated w.r.t the path variable $s$ \cite{Gao2018}). The initial and final velocity $v_0, v_\mathrm{f} \in \mathbb{R}^3$ ensure the continuity of the trajectory segments, and the inequality constraints guarantee that the velocity, acceleration, and acceleration rate remain within their allowed range. The 
bounds $v_\mathrm{max}, a_\mathrm{max}, \lambda_\mathrm{max} \in\mathbb{R}^3$ depend on the specific quadrotor configuration \TR{and} we treat them as hyperparameters. \TR{Later,}  we give an optimization-based approach to choose their values in Section~\ref{sec:kin_con}.

Optimization~\eqref{eq:temp_opt} is nonlinear and complicated to solve in this form. However, 
by the direct transcription method introduced in \cite{Verscheure2009}, it can be reformulated as a \emph{second-order cone optimization problem} (SOCP) by discretizing the temporal trajectory over a grid of $s$, and assuming that $\ddot{s}$ is piecewise constant between adjacent grid points. The numerical solution of the resulting SOCP can be obtained in polynomial time by using an off-the-shelf convex solver.

\subsection{Boundary conditions of the trajectory segments}\label{sec:const}
In this section, we formulate the corresponding boundary conditions for both the spatial and temporal trajectory of each segment, depicted in Fig.~\ref{fig:traj}. The segments are distinct in the sense that we construct both the spatial and temporal trajectory of a segment before moving on to the next one. From here on, we use the notation $r_{\mathrm{d},i}(s_i(t_i)), i\in\{1,\ldots, 5\}$ for each trajectory segment, where $s_i(0) =0$, $s_i(T_i) = 1$, and $t_i\in [0, T_i]$. 
To connect two successive trajectory segments, the continuity of the position and velocity are ensured by the following constraints:
\begin{align}
    r_{\mathrm{d},i+1}(0) = r_{\mathrm{d},_i}(1),\quad \dot{r}_{\mathrm{d},i+1}(0) =\dot r_{\mathrm{d},i}(1).
\end{align}

In the first segment of the motion sequence, the aim is to reach plane $P$ from any given initial position $r_0$ and velocity $v_0$, therefore at point $A$, the following constraints are set:
\begin{align}
    r_{\mathrm{d},1}(0) = r_0, \quad \dot r_{\mathrm{d},1}(0) = v_0.
\end{align}
At point $B$, we wish to achieve a smooth transition to plane $P$, therefore the position is required to be within the plane with zero velocity and acceleration perpendicular to the plane. To simplify the mathematical formulation of the constraints, without loss of generality, we fix the object plane $P$ to the global $x-z$ plane. In the $x$ direction, we set the position to a safety distance from the object to ensure that the controller drives the vehicle to the plane even in the presence of small tracking errors. We also constrain the direction of the spatial trajectory to ensure that the projection of the velocity vector to the $x$ axis at point $B$ points towards the payload. In the $z$ direction, we prescribe inequality constraints for both the position and the direction of the spatial trajectory to make the hook attach easier in the second segment. These constraints are formulated as follows:
\begin{align}
\begin{split}
    &y_\mathrm{d,1}(1) = 0,\quad y'_\mathrm{d,1}(1) = 0, \quad y''_\mathrm{d,1}(1) = 0,\\
    &x_\mathrm{d,1}(1) = x_B, \quad x'_\mathrm{d,1}(1) \geq \partial_s x_B,\\
    &z_{B,\mathrm{min}} \leq z_\mathrm{d,1}(1) \leq z_{B,\mathrm{max}},\\
    &\partial_s z_{B,\mathrm{min}} \leq z'_\mathrm{d,1}(1) \leq \partial_s z_{B,\mathrm{max}},
\end{split}
\end{align}
where $x_B$ and $\partial_s x_B$, $z_{B,\mathrm{min}},  z_{B,\mathrm{max}}, \partial_s z_{B,\mathrm{min}}, \partial_s z_{B,\mathrm{max}}$ are hyperparameters.

At point $C$, we prescribe constraints for the position and the direction of the spatial path to ensure that it lies in plane $P$ above the initial position of the payload, guaranteeing safe attachment of the hook:
\begin{align}
\begin{split}
        &r_{\mathrm{d},2}(1) = r_{\mathrm{L, init}} + L e_3,\\ &x'_{\mathrm{d},2}(1) \geq 0, \ \ y'_\mathrm{d,2}(1) = 0, \ \ z'_\mathrm{d,2}(1) = 0,
\end{split}
\end{align}
where $r_{\mathrm{L, init}}$ is the initial position of the payload, and $e_3$ is the unit vector of the $z$ axis in the world frame.

Between points $C$ and $D$, we transport the payload to be above the target position and stop there. We intend to achieve precise positioning at point $D$ by fixing the first and second derivatives of the spatial path to zero in $x$ and $y$ directions:
\begin{align}
\begin{split}
        &r_{\mathrm{d},3}(1) = r_{\mathrm{L, target}} + z_D,\quad \dot r_{\mathrm{d},3}(1) = 0, \\
    &x'_\mathrm{d,3}(1) = y'_\mathrm{d,3}(1) = x''_\mathrm{d,3}(1) = y''_\mathrm{d,3}(1) = 0,
\end{split}
\end{align}
where $r_{\mathrm{L, target}}$ is the target position of the payload and $z_D$ is a safety height, treated as a hyperparameter.

From point $D$ to $E$, the quadcopter descends to release the payload. As an end condition, we set the $z$ position of the drone such that the hook is detached after the end of this segment. These constraints are written as follows:
\begin{align}
    r_{\mathrm{d},4}(1) = r_{\mathrm{L, target}} + L e_3 - 0.5 d_\mathrm{h},\quad \dot r_{\mathrm{d},4}(1) = 0,
\end{align}
where $d_\mathrm{h}$ is the diameter of the hook.

Unlike all other waypoints, the spatial trajectory at point $E$ is not continuous \TR{as} Segment 5 starts with a horizontal direction to ensure that the hook is detached successfully and the payload stays on the ground. The position $r_F$ at point $F$ can be chosen arbitrarily by the user, the only goal of this trajectory segment is to make sure that the payload is released. The constraints of this segment are formulated as
\begin{align}
    r_{\mathrm{d},5}(1) = r_F,\quad z'_\mathrm{d,5}(0) = z''_\mathrm{d,5}(0) = 0.
\end{align}

Finally, we incorporate the position and spatial derivative constraints into Optimization~\eqref{eq:qp}, while the velocity and acceleration constraints are included in Optimization~\eqref{eq:temp_opt}. All boundary conditions are linear in the optimization variables, so they do not increase the complexity of the optimization.

\subsection{Yaw trajectory}\label{sec:yaw_traj}
The yaw dynamics of a quadrotor are independent of the ($x, y, z$) path, therefore the yaw reference $\psi_\mathrm{d}$ can be freely chosen. The orientation of the hook is only restricted at points $A$, $C$, $E$ in terms of the initial quadrotor orientation, the orientation of the payload (ensuring that the hook can be attached to it) and the final payload orientation.

\begin{figure}
\centering 
\includegraphics[width=.85\linewidth]{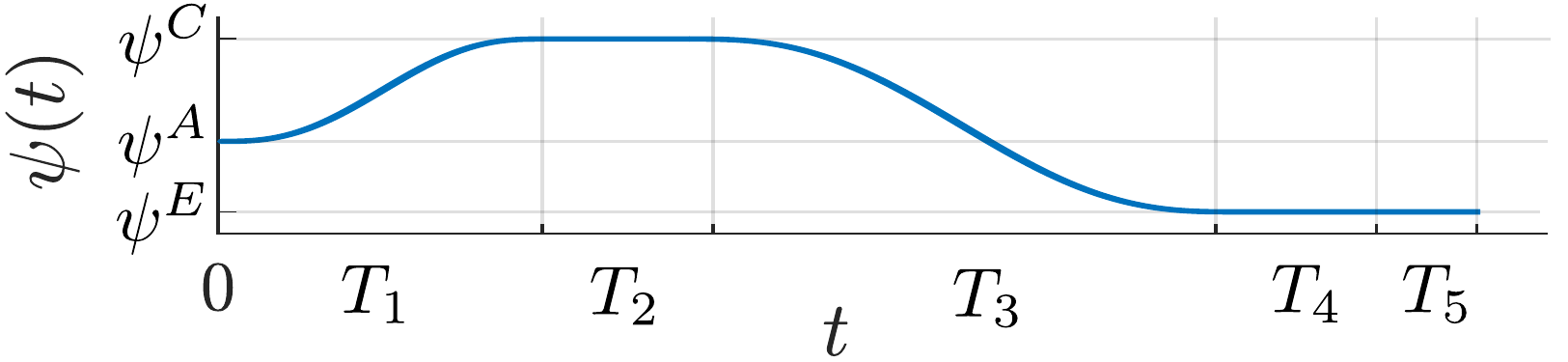}
\caption{Schematic trajectory of the yaw angle reference.}\label{fig:yaw_ref}
\end{figure}

To provide a smooth transition between the specified yaw setpoints, we design $5^\mathrm{th}$-order polynomial trajectories at Segments 1 and 3, and prescribe constant reference at the remaining segments. Given that the time duration of each segment is obtained from the solution of Optimization~\eqref{eq:temp_opt}, the initial and final values of the polynomial trajectories are known, and by prescribing zero initial and final angular velocity and angular acceleration, the coefficients of the polynomial trajectories can be directly obtained from the solution of a linear equation. A schematic trajectory is depicted in Fig.~\ref{fig:yaw_ref} to illustrate the polynomial segments.

\subsection{Hyperparameter optimization}\label{sec:kin_con}

The temporal trajectory optimization problem \eqref{eq:temp_opt} enables to prescribe constraints for the velocity, acceleration, and differential acceleration. However, exact formulation of the actuator constraints would result in a set of nonlinear equations at the acceleration level, hence convexity of the optimization would be lost. To address this issue, we treat the constraint bounds and the cost function weight in \eqref{eq:spat_obj} as hyperparameters, and propose a simple optimization program to find the best parameters that result in successful grasping and minimal execution time. By defining the hyperparameter vector as $\gamma=[\ v_\mathrm{max}\ a_\mathrm{max}\ \lambda_\mathrm{max}\ w\ ]^{\top} \in \Gamma \subset \mathbb{R}^4$, the optimization problem is formulated as
\begin{align}\label{eq:kin_con_opt}
\begin{split}
             \gamma^* =& \argmin_{\gamma\in\Gamma} \sum_{i=1}^5 T_i(\gamma),\\
    \subjto \ \ & \| r_\mathrm{b}^\mathrm{L}(T_C) \| \leq 0.5d_\mathrm{h},
\end{split}
\end{align}
where $r_\mathrm{b}^\mathrm{L}\! =\! r\! +\! Rr_\mathrm{b}^\mathrm{l}\! -\! r_\mathrm{L}$, $\|\! \cdot\! \|$ is the Euclidean vector~norm, and $d_\mathrm{h}$ is the diameter of the hook. \TR{In \eqref{eq:kin_con_opt}}, the inequality constraint ensures that the distance between the hook and the payload is less than the radius of the hook at point $C$, \TR{implying} that the grasping is successful. To set up the optimization, $\Gamma$ is chosen based on physical intuition. The cost is evaluated by drawing random initial conditions and simulating the trajectories of the system. The objective function and the constraint are difficult to express in closed form, therefore gradient-free global optimizer methods are applied to compute the solution, such as grid search, Bayesian optimization, or \emph{particle swarm optimization} (PSO). The hyperparameters are calculated offline and only once for a given quadrotor platform, thus it is feasible to use gradient-free methods despite their high computational cost.

\section{Motion Control}\label{sec:control}


\begin{figure}
    \centering
    \includegraphics[width=\linewidth]{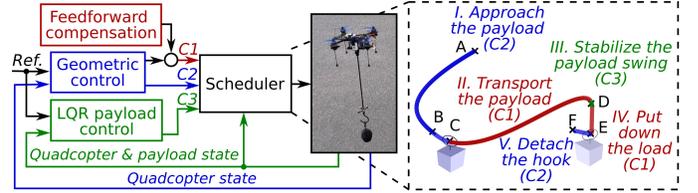}
    \caption{Block diagram of the closed-loop control scheme with scheduling of the controllers along the motion trajectory (C1: geom. with feedforward, C2: geom., C3: LQR).}
    \label{fig:control_block_diagram}
\end{figure}

\subsection{Overall concept}

We design a dual-mode motion control approach to execute the designed trajectory and ensure the successful realization of the transportation task. To follow the designed motion trajectory of the quadrotor in terms of reference position $r_\mathrm{d}$ and reference yaw angle $\psi_\mathrm{d}$, we treat the hook-based manipulator as a disturbance acting on the quadrotor dynamics and employ a robust geometric control method. Moreover, to compensate for the gravitational force of the payload, we augment the controller with a \emph{feedforward} (FF) term that is activated after attaching the hook to the payload. Robust geometric control  is able to stabilize the quadcopter under bounded uncertainty, however, it does not control the payload swing, which makes precise positioning difficult. To solve this problem, we fix the reference at point $D$ and switch to a linear--quadratic payload controller to stabilize the payload at the bottom equilibrium point. After a fixed time, the geometric controller is reactivated to finish the maneuver. The block diagram of the control scheme is displayed in Fig.~\ref{fig:control_block_diagram}. In the sequel, we introduce the control methods, and provide mathematical proof that the motion remains stable during the whole maneuver, despite the controller switching.

\subsection{Nonlinear robust geometric control}\label{sec:geom}

For precise and reliable trajectory tracking, we use a nonlinear robust geometric controller based on \cite{Lee2013}, which is able to follow a reference trajectory in SE(3) in terms of position $r_\mathrm{d}(t) = [\ x_\mathrm{d}(t)\ y_\mathrm{d}(t)\ z_\mathrm{d}(t)\ ]^\top \in \mathbb{R}^3$ and orientation $R_\mathrm{d}(t) \in \mathrm{SO}(3)$, where SE(3) denotes the special Euclidean group that comprises arbitrary translations and rotations of a rigid body in 3D space. The reference orientation $R_\mathrm{d}(t)$ is fully determined by the reference position $r_\mathrm{d}(t)$ and yaw angle $\psi_\mathrm{d}(t)$. To synthesize the control law, the quadrotor dynamics given by \eqref{eq:quad_dyn} are augmented by additive state-dependent disturbances:
\begin{subequations}\label{eq:unc_quad_dyn}
\begin{align}
    m\ddot{r} &= -m g e_3 + F R e_3 + \Delta_\mathrm r(\state),\\
    J\dot{\omega} &= \tau - \omega \times J \omega + \Delta_\mathrm R(\state),
\end{align}
\end{subequations}
where $\state = [\ \xi^\top\ \dot\xi^\top\ ]^\top$, and $\Delta_\mathrm r(\state), \Delta_\mathrm R(\state) \in \mathbb{R}^3$ comprise the disturbance caused by the payload motion and the hook-based manipulator on the quadrotor, bounded by $ \| \Delta_\mathrm r(\state) \| \leq \delta_\mathrm r$, $\| \Delta_\mathrm R(\state) \| \leq \delta_\mathrm R$, where $\delta_\mathrm r, \delta_\mathrm R \in \mathbb{R}$ are fixed constants. Force and torque inputs are computed based on the following control law:
\begin{subequations}\label{eq:geomlaw}
    \begin{align}
        F =&  (-k_\mathrm{r}e_\mathrm{r} - k_\mathrm{v}e_\mathrm{v} + mge_3 + m\ddot{r}_\mathrm{d} + \mu_\mathrm{r})^\top R e_3,\label{eq:geomforce}\\
        \begin{split}
            \tau =&  -k_\mathrm{R} e_\mathrm{R} - k_\omega e_\omega + \omega \times J\omega \\ &-J\left(\hat{\omega} R^\top R_{\mathrm{d}} \omega_{\mathrm{d}}-R^\top R_{\mathrm{d}} \dot{\omega}_{\mathrm{d}} + \mu_\mathrm{R}\right),
        \end{split}\label{eq:geomtau}
      \end{align}
\end{subequations}
where $k_\mathrm{r}, k_\mathrm{v}, k_\mathrm{R}, k_\omega \in \mathbb{R}$ are the control gains. The tracking errors $e_\mathrm{r}(t), e_\mathrm{v}(t), e_\mathrm{R}(t), e_\omega(t) \in \mathbb{R}^3$ and the robust control inputs $\mu_\mathrm{r}(t), \mu_\mathrm{R}(t)\in \mathbb{R}^3$ are defined as follows:
\begin{align}
    e_\mathrm{r} &= r - r_\mathrm{d}, &
    e_\mathrm{v} &= \dot{r} - \dot{r}_\mathrm{d}.\nonumber\\
    e_\mathrm{R} &= \frac{1}{2}\left(R_\mathrm{d}^\top R - R^\top R_\mathrm{d}\right)^\vee, &
    e_\omega &= \omega-R^\top R_\mathrm{d}\omega_\mathrm{d},\nonumber\\
    \mu_{\mathrm{r}} &=-\frac{\delta_{\mathrm{r}}^{\kappa+2} e_{\mathrm{B}}\left\|e_{\mathrm{B}}\right\|^{\kappa}}{\delta_{\mathrm{r}}^{\kappa+1}\left\|e_{\mathrm{B}}\right\|^{\kappa+1}+\epsilon_{\mathrm{r}}^{\kappa+1}}, &
    e_{\mathrm{B}} &=e_{\mathrm{v}}+\frac{c_{1}}{m} e_{\mathrm{r}}, \label{eq:robust_terms}\\
    \mu_{\mathrm{R}} &=-\frac{\delta_{\mathrm{R}}^{2} e_{\mathrm{A}}}{\delta_{\mathrm{R}}\left\|e_{\mathrm{A}}\right\|+\epsilon_{\mathrm{R}}}, &
    e_{\mathrm{A}} &=e_{\omega}+c_{2} J^{-1} e_{\mathrm{R}},\nonumber
\end{align}  
where $ c_{1}, c_{2}, \epsilon_{\mathrm{r}}, \epsilon_{\mathrm{R}}, \kappa$ are positive constants, $\kappa>2$, and the \emph{vee operator} $(\cdot)^\vee:\mathrm{SO}(3)\rightarrow \mathbb{R}^3$ is the inverse of the hat operator $\hat{(\cdot)}$ introduced in Section~\ref{sec:model}. It is proven in \cite{Lee2013} that the tracking errors $(e_\mathrm{r}, e_\mathrm{v}, e_\mathrm{R}, e_\omega)$ are uniformly ultimately bounded, using the control law \eqref{eq:geomlaw}, when the control gains are chosen according to the stability conditions in \cite{Lee2013}.

To obtain the uncertainty bounds $\delta_\mathrm r, \delta_\mathrm R$, we maximize the norm of the difference between the full manipulator model \eqref{eq:full_dyn} and the quadcopter model \eqref{eq:quad_dyn} over the operating domain. For this, we transform \eqref{eq:full_dyn} using $\omega = Q\dot\lambda$, as follows:
\begin{subequations}
    \begin{align}
     m\ddot r  &= \underbrace{m\left[\arraycolsep=2pt\def\arraystretch{0.8}
\begin{array}{*{14}c}
        I & 0_{3\times 3} & 0_{3\times 1}
    \end{array}\right] f_\mathrm{full}(\state, u)}_{ f_\mathrm{full}^\mathrm{r}},\label{eq:full_nl_1} \\
     J\dot\omega &= \underbrace{J\left[\arraycolsep=2pt\def\arraystretch{0.8}
\begin{array}{*{14}c} 0_{3\times 3}& Q & 0_{3\times 1} \end{array}\right] f_\mathrm{full}(\state, u)  + \dot Q Q^{-1} \omega}_{f_\mathrm{full}^\omega},\label{eq:full_nl_2}
\end{align}
\end{subequations}
where $f_\mathrm{full}(\state, u)$ is expressed by rearranging \eqref{eq:full_dyn} such that $\ddot\xi = f_\mathrm{full}(\state, u)$. Then, we formulate the following optimization problems to find $\delta_\mathrm{r}, \delta_\mathrm{R}$:
\begin{subequations}\label{eq:find_deltas}
\begin{align}
       \delta_\mathrm{r} &= \max_{\substack{\state\in \mathcal{X} \\u \in \mathcal{U} }} \left\{ \left\Vert  f_\mathrm{full}^\mathrm{r} - f_\mathrm{quad}^\mathrm{r} \right\Vert \right\},\\
     \delta_\mathrm{R} &= \max_{\substack{\state\in \mathcal{X} \\u \in \mathcal{U} }} \left\{ \left\Vert  f_\mathrm{full}^\omega- f_\mathrm{quad}^\omega \right\Vert \right\},
\end{align}
\end{subequations}
where $\mathcal{X}$ is the operating region and $\mathcal{U}$ is the set of admissible control inputs. Optimization~\eqref{eq:find_deltas} is solved using CasADi with IPOPT, which is an efficient nonlinear optimizer \cite{Andersson2019}.

Before payload grasping, i.e., between points $A$ and $C$ of the trajectory illustrated in Fig.~\ref{fig:traj}, we use the control law \eqref{eq:geomlaw}. However, the payload changes the mass of the system significantly, therefore we augment the controller with a feedforward thrust to compensate for the gravitational force:
\begin{align}\label{eq:ffthrust}
    F_\mathrm{L} = F + m_\mathrm{L} g e_3^\top R e_3 = F+ F_\mathrm{ff}.
\end{align}
The resulting thrust given by \eqref{eq:ffthrust} is able to compensate for the vertical tracking error imposed by the transported payload, if the orientation of the quadrotor is near hovering, otherwise the robustness of the control law ensures that the quadcopter body remains stable under the bounded disturbance caused by the swinging of the payload.

\subsection{Linear--quadratic payload regulator}\label{sec:lqr}

The geometric controller renders the quadcopter able to perform fast, agile maneuvers, however, it does not control the payload swing, which makes precise positioning of the payload difficult. To make sure that the payload is stabilized at the equlibrium point before detaching the hook, we \TR{use} a \emph{linear--quadratic regulator} (LQR) \cite{Kwakernaak}. 

We formulate the local LTI dynamics based on the first-order Taylor approximation of the nonlinear equations of motion described by \eqref{eq:full_dyn} as follows:
\begin{align}\label{eq:linearize}
    \dot\state &\approx \frac{\partial f(\state, u)}{\partial \state}\bigg\vert_{\state_0,u_0} \!\!\! (\state - \state_0) + \frac{\partial f(\state, u)}{\partial u}\bigg\vert_{\state_0,u_0} \!\!\! (u - u_0),
\end{align}
where $\state = [\ \xi^\top\ \dot\xi^\top \ ]^\top$, $u = [\ F\ \tau^\top\ ]^\top$, $\state_0 = 0_{14\times 1}$, $u_0 = [\ (m + m_\mathrm{L}) g \ 0_{1\times 3} \ ]^\top$, and $f = [\ \dot\xi^\top \ f_\mathrm{full}(\state, u)^\top]^\top$. 
The control input provided by the LQR controller is calculated as $u = u_0 - K (\state - \state_0)$, 
where $K\in \mathbb{R}^{4 \times 14}$ is the feedback gain matrix that minimizes the cost function $\int_0^\infty \left((\state-\state_0)^\top W_\mathrm{Q} (\state-\state_0) + (u - u_0)^\top W_\mathrm{R} (u-u_0)\right) \mathrm{d} t$ with respect to the LTI dynamics described by \eqref{eq:linearize}.

\subsection{Stability analysis of controller switching}

During the execution of the transportation task, two controller switches occur: from geometric control to LQR after Segment~3 and back to geometric control before Segment~4. In this subsection, we analyze the stability of the closed-loop system at these two switching instants, respectively.

Stability analysis of nonlinear systems is often performed by Lyapunov's method \cite{Khalil2002}, however, constructing a Lyapunov function for a high dimensional system is challenging in general. Hence, we employ a data-driven technique based on the scenario approach \cite{Campi2018} to analyze the stability of the closed-loop system $f_\mathrm{cl}(\state) = f_\mathrm{full}(\state, u_0\! -\! K(\state\! -\!\state_0))$ governed by the LQR controller. For this, first we select the operating region of the system as follows: $\mathcal{X} = \{\state \ | \ | \state_i - \state_{0,i} | \leq a_i,\ a_i \in \mathbb{R}^+,\ i=1,\dots,14 \}$, where $a$ contains the bounds on the absolute value of each element of $\state-\state_0$. Then, we intend to prove that any trajectory starting in $\mathcal{X}$ converges to zero, which certifies that $\mathcal{X}$ is a subset of the \emph{region of attraction} (ROA) of the nonlinear system \cite{Khalil2002}.


To prove our hypothesis, $N$ number of independent, identically distributed samples are drawn from $\mathcal{X}$, considered as initial states, and the convergence of the corresponding trajectories are checked in simulation. Using the terminology of \cite{Campi2018}, each simulation is interpreted as a \textit{scenario}, and the convergence of the corresponding trajectory is the \textit{decision} $\theta \in \Theta$, $\Theta = \{0, 1\}$, where 1 corresponds to stability and 0 to instability. Hence, the \textit{scenario decision} $\theta_N^*$ is also either 1 if all trajectories converge or 0 otherwise. The \textit{violation probability} of a given decision $\theta\in\Theta$ is defined as $\mathsf{V}(\theta):=\mathsf{P}\{\state^0\in\mathcal{X}:\theta = 0 \}$, where $\state^0$ is a sampled initial state, and $\mathsf{P}$ denotes probability. In \cite{Campi2018}, it is proven that, given a \textit{confidence parameter} $\beta$ and number of drawn samples $N$, \TR{it holds that} $\mathsf{P}\{\mathsf{V}(\theta_N^*) > \varepsilon(s_N^*)\} \leq \beta$, where $s_N^*$ is the \textit{cardinality} of the scenarios, while $\varepsilon(k) = 1-\sqrt[N-1]{\beta N^{-1} \binom{N}{k}^{-1}}$ for $k\in \mathbb{Z}, 1 \leq k < N$ and $\varepsilon(N)=1$. In general, $s_N^*$ and thus $\varepsilon(s_N^*)$ cannot be computed a priori, because \TR{they} depend on the sample points. However, in our case, $s_N^*=1$ because if all trajectories converge, then the scenario decision is $\theta_N^*=1$, while a single divergence supports $\theta_N^*=0$. Hence, the general results of \cite{Campi2018} can be simplified, as follows:
\begin{subequations}
    \begin{align}
    &\mathsf{P}\{\mathsf{V}(\theta_N^*) > \varepsilon(1)\} \leq \beta,\label{eq:scenario_a}\\
    &\varepsilon (1) = 1 - \sqrt[N-1]{\beta N^{-2}}.\label{eq:scenario_b}
\end{align}
\end{subequations}
Based on \eqref{eq:scenario_b}, it is possible to compute the minimum number of samples to satisfy \eqref{eq:scenario_a} for given $\beta$ and $\varepsilon(1)$. For example, \TR{if} we select $\beta=10^{-6}$ and $\varepsilon(1) = 0.001$, then $N=35000$ by \eqref{eq:scenario_b}. If this many samples are drawn, and all trajectories converge, then, based on \cite{Campi2018}, $\mathcal{X}$ is proven to be a subset of the ROA with $1-\varepsilon(1) = 99.9 \%$ probability. 

After the setpoint stabilization of the system at~point~$D$, the geometric controller is reactivated to finish the transportation task by detaching the hook. In this case, the ROA of the closed-loop system is described in \cite{Lee2013} in terms of the physical parameters and the control gains. Therefore at the beginning of Segment 4, we check the conditions of \cite{Lee2013} to obtain whether the closed loop is stable and the switching can be executed safely.


\begin{table}
\centering
     \caption{Parameters of the manipulator and the geometric controller.} 
     \label{tab:params}\begin{tabular}{lcrlcc}
    \hline
        Drone mass & $m$ & $6.05\cdot 10^{-1}$ & kg & $k_\mathrm{r}$ & 6.0\\
        $x$ axis inertia &  $J_\mathrm{x}$ & $1.5\cdot 10^{-3}$ &kgm$^2$ & $k_\mathrm{v}$ & 3.0\\
        $y$ axis inertia & $J_\mathrm{y}$ & $1.45\cdot 10^{-3}$ &kgm$^2$ & $k_\mathrm{R}$ & 1.0\\
        $z$ axis inertia &  $J_\mathrm{z}$ &  $2.66\cdot 10^{-3}$ &kgm$^2$ & $k_\omega$ & 0.2\\
        Rotor distance & $l$ & $8.3\cdot 10^{-2}$ & m & $c_1$ & 1.0\\
        Pole length & $L$ & $4.0\cdot 10^{-1}$ & m & $c_2$ & $10^{-3}$ \\
        Hook mass & $m_\mathrm{h}$ & $1.0\cdot 10^{-2}$ & kg  & $\epsilon_\mathrm{r}, \epsilon_\mathrm{R}$ & $10^{-4}$\\
        Hook diameter & $d_\mathrm{h}$ & $4.0\cdot 10^{-2}$ & m  & $\kappa$ & 3\\
        \hline
    \end{tabular}
\end{table}

\section{Simulation Study}\label{sec:simulation}

\subsection{Simulation environment}

\begin{figure}
    \centering
    \includegraphics[height=4.3cm]{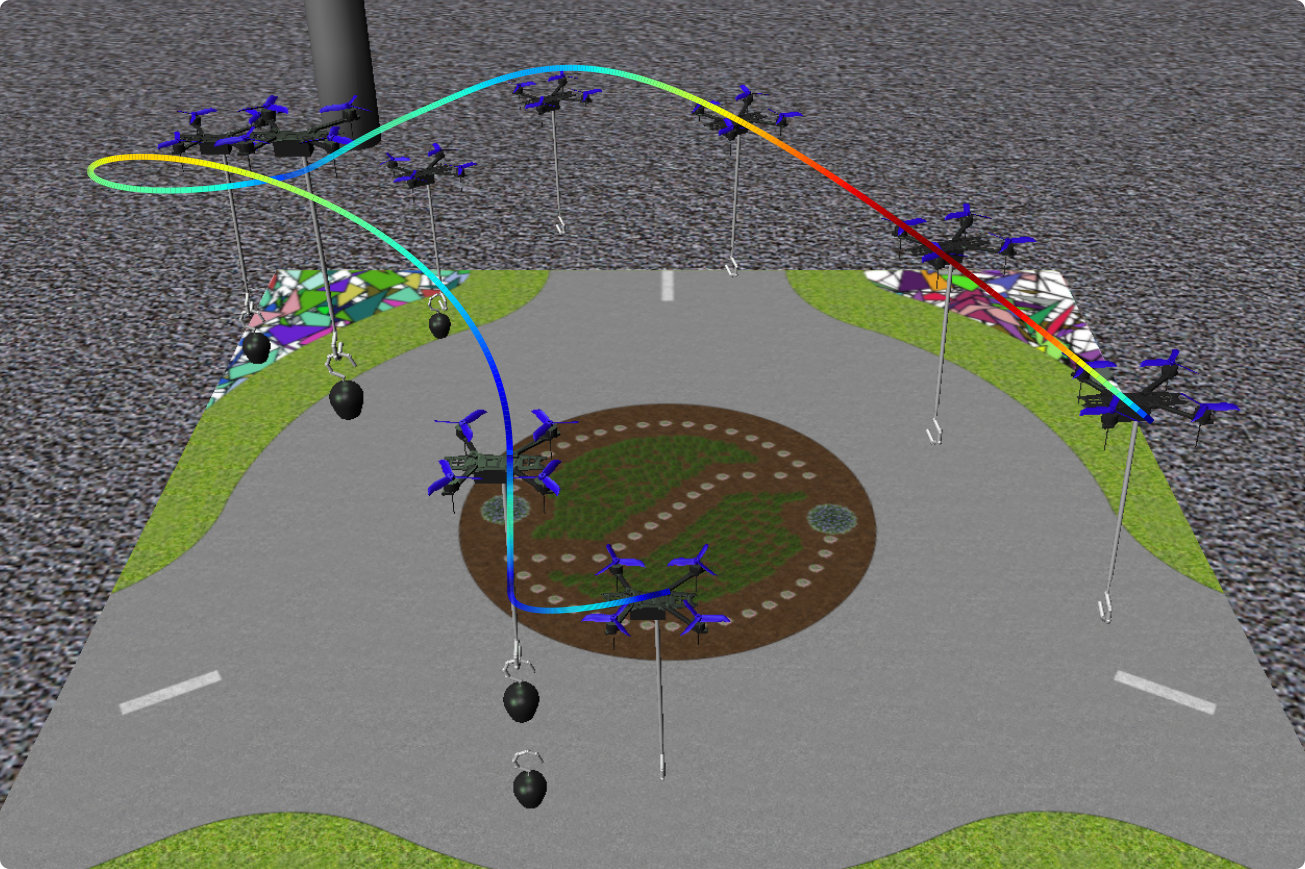}
    \caption{Simulated transportation scenario using the digital twin model of the hook-based manipulator and the test environment. Note the close resemblance to Fig.~\ref{fig:real_manipulator}, where the real-world experiments are illustrated.}
    \label{fig:composite_simu}
\end{figure}

\TR{First,} we analyze the proposed trajectory planning and control methods in a high-fidelity physics simulator in emulated real-world scenarios. The physical parameters of our custom-made quadrotor--hook manipulator platform are shown in Table~\ref{tab:params}. To construct realistic simulations, we have implemented the digital twin model of the manipulator and the test environment, illustrated in Fig.~\ref{fig:composite_simu}. The dynamics are simulated by MuJoCo \cite{todorov2012mujoco}, an open-source physics engine that makes it possible to efficiently simulate complex dynamic systems with collisions and contacts. 
The source code of the simulations is available at GitHub\footnote{\url{https://github.com/AIMotionLab-SZTAKI/quadcopter\_hook\_1DoF}}.

\subsection{Trajectory planner design}

To optimize the hyperparameters of the trajectory planner, \eqref{eq:kin_con_opt} has been solved by grid search on the region $\Gamma=\{ \gamma \ |\ v_\mathrm{max} \in [0.7, 3]\ \mathrm{m/s}$, $a_\mathrm{max} \in [0.1, 0.5]\ \mathrm{m/s}^2$, $\lambda_\mathrm{max} \in [0, 0.2]$, $w \in [0.01, 0.9]$\}, in which 10 uniformly distributed values for $v_\mathrm{max}, a_\mathrm{max}, \lambda_\mathrm{max}$ and 4 values for $w$ have been considered as grid points (4000 cases). To test the feasibility and execution time of each parameter combination, 6 different scenarios of the grasping maneuver have been simulated in terms of the initial and final configuration of the hook-based manipulator. The obtained numerical values are $v_\mathrm{max}^* = 1.46\ \mathrm{m/s}$, $a_\mathrm{max}^* = 0.2\ \mathrm{m/s}^2$, $\lambda_\mathrm{max}^* = 0.1$, $w^* = 0.5$.

Besides the optimized variables, there are other parameters, namely the degree $n_\mathrm{r}$ and the number of coefficients $n_\mathrm{c}$ of the B-spline curve representing the spatial trajectory, the number of grid points $K$ by the temporal trajectory, and the control input weight $\rho$ in Optimization \eqref{eq:temp_opt}. The numerical values of these parameters have been chosen to be $n_\mathrm{r} = 5, n_\mathrm{c} = 12, K = 15, \rho = 0.1$ based on the guidelines derived in \cite{Gao2018} for similar trajectory optimization problems.



For each of the 5 segments detailed in Sec.~\ref{sec:overview}, a \emph{quadratic program} (QP) has been constructed for the spatial trajectory, and a \emph{second-order cone program} (SOCP) has been set up for the temporal trajectory. To solve these problems, we have used Mosek\footnote{\url{https://www.mosek.com/}}, a large scale, efficient optimization toolbox. On a Windows notebook with Intel i5 CPU and 8GB RAM, the average computation time of the QP is $T_\mathrm{QP} = 3.30$ ms, while, for the SOCP, is $T_\mathrm{SOCP} = 42.7$ ms. These values correspond to one of the five segments, therefore the full optimization time for grasping, transportation and release is $T_\mathrm{opt,full} = 5(T_\mathrm{QP} + T_\mathrm{SOCP}) = 230$ ms in average. However, the construction of constraint matrices and other matrix operations are computationally demanding, therefore the full computation time of the trajectory planning is significantly larger than the pure optimization times, it is $T_\mathrm{full} = 1.54$~s in average. Although this computation time is already acceptable for real-world applications, it can be greatly reduced by embedded implementation.

\subsection{Motion control design}

The numerical values of the parameters of the robust geometric control are provided in Table~\ref{tab:params}, which we have chosen based on the physical parameters of the quadcopter platform and the stability conditions of \cite{Lee2013}. By solving Optimization~\eqref{eq:find_deltas} with these parameters, the result yields $\delta_r=0.28$ N, $\delta_R = 0.011$ Nm. These values are similar to the disturbance magnitudes considered in \cite{Lee2013} relative to the physical properties of the quadcopter platform.

The weights of the LQR controller and the bounds of the operating region $\mathcal{X}$ have been chosen, as follows: $W_\mathrm{Q} = \mathrm{diag}(10, 10, 100, 0.1, 0.1, 0.1, 0.01, 1, 1, 10, 1, 1, 1, 0.05),$ $W_\mathrm{R} = \mathrm{diag}(5, 20, 20, 20),\ a_i = 0.2$ for $i\in \{ 7, 14 \}$, and $a_i = 0.1$ otherwise. 
For the stability analysis, $N=35000$ uniformly distributed samples have been generated within $\mathcal{X}$. All simulated trajectories have converged, therefore under $\beta=10^{-6}$, we conclude that with $99.9 \%$ certainty the closed-loop system is stable after controller switching, given that the geometric controller brings all states within $\mathcal{X}$ by the end of Segment~3. The latter condition is satisfied in all of the presented simulations, and in the real-world experiments, as well.

\begin{figure*}
\centering 
\includegraphics[width=.97\linewidth]{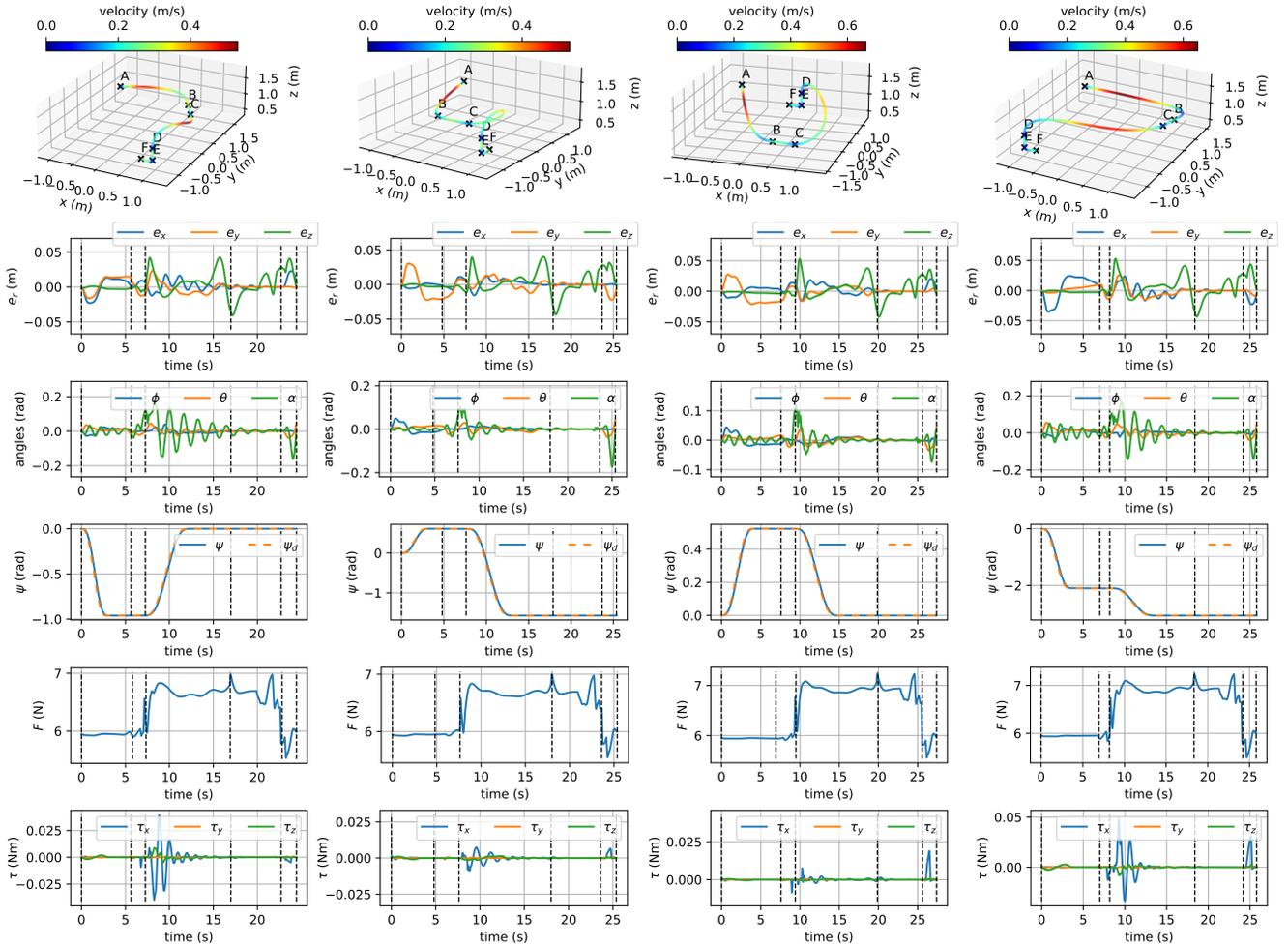}
\caption{Simulated scenarios of payload transportation: the 3D plots display the reference trajectories, while the 2D plots show the state and error trajectories. The payload mass has been set to 75 g at the first and second scenarios, and to 100 g at the third and fourth. The waypoints $A, B, C, D, E, F$ are indicated by dashed lines.}\label{fig:simu}
\end{figure*}


\subsection{Results}

To challenge the proposed methods, four transportation scenarios have been simulated in MuJoCo. The results are shown in Fig.~\ref{fig:simu}, where each column corresponds to a specific scenario. The first row shows the reference position and velocity, proving that smooth trajectories can be generated using the proposed planning method to rapidly grasp the payload, take it to the target location, and release it. 

To evaluate the effect of different payloads, the payload mass has been set to 75~g at the first two scenarios, and 100~g at the other two simulations. The root mean square error of the position tracking for each scenario has been $e_\mathrm{r, RMSE} = \{2.1, 2.1, 2.1, 2.3\}$~cm, the peak absolute error has been $e_\mathrm{r, max} = \{5.5, 5.3, 5.9, 5.9\}$~cm, and the final position error of each payload has been less than 1~mm. The trajectory of the swing angle $\alpha$ shows a maximum of 0.18~rad that appears after grasping the last payload. After attaching the hook, the swing of each payload becomes significant, however, the LQR is able to damp payload oscillations in all cases to achieve precise positioning when detaching the hook. The yaw angle tracking is also accurate, the difference between the reference and state trajectories are insignificant. To conclude, the simulations show excellent control performance both in terms of reference tracking and robustness against the variation of the payload mass.

\section{Flight Experiments}\label{sec:exp}



\begin{figure*}
\centering 
\includegraphics[width=\linewidth]{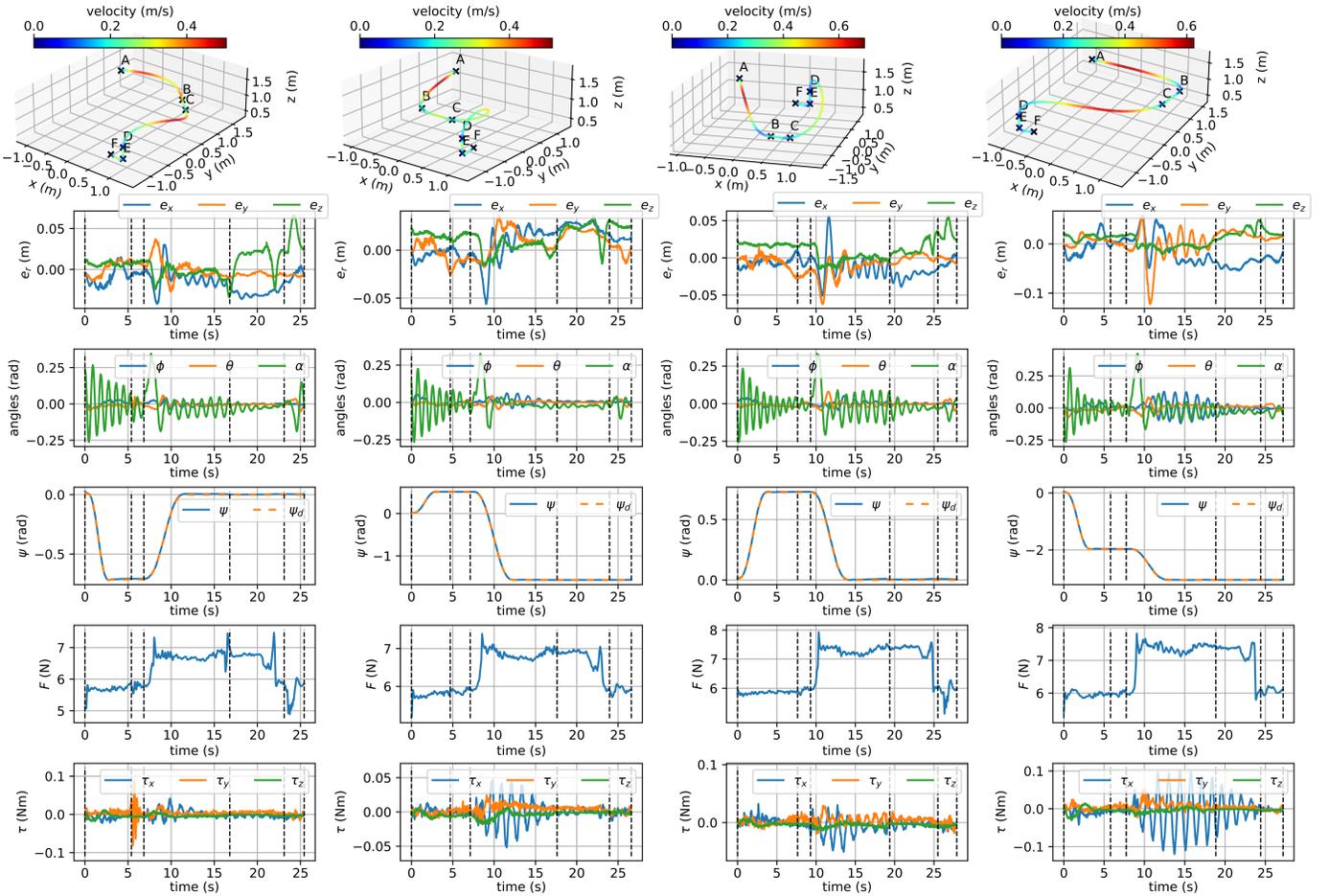}
\caption{Real-world transportation experiment: the reference trajectories, states, inputs, and waypoints are indicated identically to the simulation results.}\label{fig:meas}
\end{figure*}

\subsection{Experimental setup}

Flight experiments have been conducted with the proposed method, implemented on a custom-made quadrotor platform equipped with a 3D printed hook and revolute joint, shown in Fig.~\ref{fig:real_manipulator}. The on-board computation (including state estimation and control input calculation at 500 Hz) runs on a {Crazyflie Bolt flight controller}\footnote{\url{https://www.bitcraze.io/products/crazyflie-bolt-1-1/}}, including an IMU and two microcontroller units. Optitrack motion capture system is used to provide high-precision pose information. A more detailed specification of the experimental setup is available in \cite{Antal2022_2}. The physical parameters of the hook-based manipulator are shown in Table~\ref{tab:params}, and for the flight experiments, similarly to the simulation study, we use a 75 g and a 100 g payload.

\changes{In real-world experiments, only the target position and orientation of the payload needs to be specified by the user. The trajectory planner then automatically detects the pose of the quadcopter and the payload based on the motion capture system, runs the planning algorithm, and sends the trajectory to the vehicle. After the transportation is completed, a descending trajectory is computed and sent to the quadcopter. This way, the \TR{quadcopter} is able to safely land with the attached manipulator.}

\subsection{Results}

Measurement results for the flight experiments are displayed in Fig.~\ref{fig:meas}, where the reference trajectories, states, and control inputs are shown similarly to the simulations in Fig.~\ref{fig:simu}. The root mean square error of the position tracking for each scenario has been $e_\mathrm{r, RMSE} = \{2.8, 2.9, 2.9, 4.3\}$~cm, and the final position error of each payload is $e_\mathrm{r, fin} = \{1.8, 1.7, 1.0, 3.1\}$~cm, respectively. The trajectory of the swing angle $\alpha$ shows a maximum of 0.34~rad that appears after grasping the last payload with the highest mass. However, it is also shown experimentally, that the LQR is able to damp payload oscillations to achieve precise positioning when detaching the hook. Similar to the simulations, the yaw angle tracking error is insignificant in real experiments as well. Based on the measurement results, we conclude that the control performance is robust against the payload mass variation, and it is excellent in all examined scenarios, therefore the proposed methods have been validated successfully in real-world deployment.


It is important to note that unmodeled aerodynamics, more specifically drag, downwash and ground effect can have a significant impact on the behaviour of the system in real flight, see \cite{Bauersfeld2021,Powers2013}. For flat and light objects that have large aerodynamic resistance, the \TR{airflow} generated by the propellers can cause the payload to swing on the hook, which can be challenging to compensate by the controller. Therefore, the modelling of this aerodynamic effect and \TR{compensation of it} in the control design is an important task for future research.

Video footage of the flight experiments is available at \url{https://youtu.be/FDqSrf4k9kw}.

\section{Conclusion}\label{sec:summary}

In this work, we have presented a complete methodology for payload grasping, transportation, and release using a quadrotor with a hook-based manipulator. To validate the proposed method, 
we have deployed the algorithms on a hardware platform and performed flight experiments. The demonstrated results show that the proposed trajectory planning and control methods are time efficient, robust, and can be applied in real flights.  


\changes{
\section*{Acknowledgments}
The authors would like to thank Botond Gaál for his help during the flight experiments.}

{\footnotesize
\bibliography{reference}
}

\end{document}